\documentclass{article}
\usepackage{hyperref}
\usepackage{nips14submit_e,times}
\usepackage{graphicx}
\usepackage[numbers]{natbib}
\usepackage{algorithm,algorithmic}

\usepackage{tikz}
\usetikzlibrary{calc}
\def\tikzmark#1{\tikz[remember picture,overlay]\coordinate(#1);}

\usepackage{macros}

\def\x{\vx}
\def\i{^{(i)}}
\def\xopt{\vx_\star}
\def\xrec{\widetilde\vx}
\def\X{\calX}
\def\D{\calD}
\def\H{\mathrm{H}}

\title{Predictive Entropy Search for Efficient Global Optimization of Black-box
Functions}

\author{
Jos\'e Miguel Hern\'andez-Lobato\\
\texttt{jmh233@cam.ac.uk}\\
University of Cambridge
\And
Matthew W. Hoffman\\
\texttt{mwh30@cam.ac.uk}\\
University of Cambridge
\And
Zoubin Ghahramani\\
\texttt{zoubin@eng.cam.ac.uk}\\
University of Cambridge
}

\nipsfinalcopy

\begin{document}
\maketitle

\begin{abstract}
We propose a novel information-theoretic approach for Bayesian optimization
called Predictive Entropy Search (PES). At each iteration, PES selects the next
evaluation point that maximizes the expected information gained with respect to
the global maximum. PES codifies this intractable acquisition function in terms
of the expected reduction in the differential entropy of the predictive
distribution.  This reformulation allows PES to obtain approximations that are
both more accurate and efficient than other alternatives such as Entropy Search
(ES). Furthermore, PES can easily perform a fully Bayesian treatment of the model
hyperparameters while ES cannot. We evaluate PES in both synthetic and
real-world applications, including optimization problems in machine learning,
finance, biotechnology, and robotics. We show that the increased accuracy of PES
leads to significant gains in optimization performance.
\end{abstract}


\section{Introduction}

Bayesian optimization techniques form a successful 
approach for optimizing black-box functions \cite{Brochu:2009}. The goal
of these methods is to find the global maximizer of a nonlinear and generally non-convex function $f$
whose derivatives are unavailable. Furthermore, the evaluations of $f$ are usually corrupted by noise and
the process that queries $f$ can be computationally or economically very expensive. To address these challenges,
Bayesian optimization devotes additional effort to modeling the unknown function $f$ and its behavior.
These additional computations aim to minimize the number of evaluations that are needed to find the global optima.

Optimization problems are widespread in science and engineering and as a result
so are Bayesian approaches to this problem.  Bayesian optimization has
successfully been used in robotics to adjust the parameters of a robot's
controller to maximize gait speed and smoothness~\citep{Lizotte:2007} as well as
parameter tuning for computer graphics~\cite{Brochu:2007}. Another example
application in drug discovery is to find the chemical derivative of a particular
molecule that best treats a given disease~\cite{negoescu:2011}.  Finally,
Bayesian optimization can also be used to find optimal hyper-parameter values
for statistical~\cite{Wang:2013} and machine learning techniques
\citep{Snoek:2012}.

As described above, we are interested in finding the global maximizer $\xopt=\argmax_{\x\in\X} f(\x)$
of a function $f$ over some bounded domain, typically $\X\subset\R^d$. We 
assume that $f(\x)$ can only be evaluated via queries to a black-box that
provides noisy outputs of the form $y_i\sim \Normal(f(\x_i), \sigma^2)$. We
note, however, that our framework can be extended to other non-Gaussian
likelihoods. In this setting, we describe a sequential search algorithm that,
after $n$ iterations, proposes to evaluate $f$ at some location $\x_{n+1}$. To
make this decision the algorithm conditions on all previous observations
$\D_n=\{(\x_1,y_1),\dots,(\x_n,y_n)\}$. After $N$ iterations the algorithm 
makes a final recommendation $\xrec_N$ for the global maximizer of the latent function $f$.  

We take a Bayesian approach to the problem described above and use a probabilistic model for the
latent function $f$ to guide the search and to select $\xrec_N$.
In this work we use a zero-mean Gaussian process (GP) prior for $f$
\cite{Rasmussen:2006}. This prior is specified by a positive-definite kernel
function $k(\x,\x')$. Given any finite collection of points
$\{\x_1,\dots,\x_n\}$, the values of $f$ at these points are jointly zero-mean
Gaussian with covariance matrix $\vK_n$, where $[\vK_n]_{ij}=k(\x_i,\x_j)$. For
the Gaussian likelihood described above, the vector of concatenated observations $\vy_n$
is also jointly Gaussian with zero-mean. Therefore, at any
location $\x$, the latent function $f(\x)$ conditioned on past observations
$\D_n$ is then Gaussian with marginal mean $\mu_n(\x)$ and variance $v_n(\x)$
given by
\begin{align}
    \mu_n(\x) 
    &= \vk_n(\x)\T(\vK_n+\sigma^2\vI)^{-1}\vy_n\,, &
    v_n(\x)
    &= k(\x,\x) - \vk_n(\x)\T(\vK_n+\sigma^2\vI)^{-1}\vk_n(\x)\,,
    \label{eq:gpvar}
\end{align}
where $\vk_n(\x)$ is a vector of cross-covariance terms between $\x$ and
$\{\x_1,\dots,\x_n\}$.

Bayesian optimization techniques use the above predictive distribution $p(f(\x)|\D_n)$
to guide the search for the global maximizer $\xopt$. In particular, 
$p(f(\x)|\D_n)$ is used during the computation of an acquisition function $\alpha_n(\x)$
that is optimized at each iteration to determine the next evaluation location $\x_{n+1}$.
This process is shown in Algorithm~\ref{alg:bo}. Intuitively, the acquisition function
$\alpha_n(\x)$ should be high in areas where the maxima is most likely to lie given the current data.
However, $\alpha_n(\x)$ should also encourage exploration of the search space to
guarantee that the recommendation $\xrec_N$ is a global optimum of $f$, not just a global
optimum of the posterior mean.  Several acquisition functions have been proposed
in the literature. Some examples are the probability of
improvement~\citep{Kushner:1964}, the expected improvement~\citep{Mockus:1978}
or upper confidence bounds~\citep{Srinivas:2010}.  Alternatively, one can
combine several of these acquisition functions~\citep{Hoffman:2011}.

\begin{figure}
\vspace{-1.5em}
\setlength{\tabcolsep}{0pt}
\begin{tabular}{p{0.5\textwidth} p{0.5\textwidth}}
\vspace{0pt}
\begin{minipage}{0.48\textwidth}
\begin{algorithm}[H]
    \caption{Generic Bayesian optimization}
    \label{alg:bo}
    {\footnotesize
    \begin{algorithmic}[1]
        \REQUIRE a black-box with unknown mean $f$
        \FOR{$n=1,\dots,N$}
            \STATE select $\x_n = \argmax_{\x\in\X}\alpha_{n-1}(\x)$
            \STATE query the black-box at $\x_n$ to obtain $y_n$
            \STATE augment data $\D_n = \D_{n-1} \cup \{(\x_n, y_n)\}$
        \ENDFOR
        \RETURN $\xrec_N=\argmax_{\x\in\X} \mu_N(\x)$
        \vspace{3.2em}
    \end{algorithmic}}
    \end{algorithm}
\end{minipage}
&
\vspace{0pt}
\begin{minipage}{0.48\textwidth}
    \begin{algorithm}[H]
    \caption{PES acquisition function}
    \label{alg:pes}
    {\footnotesize
    \begin{algorithmic}[1]
        \REQUIRE a candidate $\x$; data $\D_n$
        \STATE \tikzmark{a}sample $M$ hyperparameter values $\{\vpsi\i\}$
        \FOR{$i=1,\dots,M$}
            \STATE sample $f\i\sim p(f|\D_n, \vphi, \vpsi\i)$
            \STATE set $\xopt\i\gets\argmax_{\x\in\X} f\i(\x)$
            \STATE compute $\vm_0\i$, $\vV_0\i$ and $\widetilde\vm\i$, $\widetilde\vv\i$
            \STATE compute $v_n\i(\x)$ and $v_n\i(\x|\xopt\i)$
        \ENDFOR
        \RETURN $\alpha_n(\x)$ as in (\ref{eq:marginalizedObjective})
    \end{algorithmic}}
    \end{algorithm}
\end{minipage}
\end{tabular}
\begin{tikzpicture}[remember picture, overlay]
    \node
    [draw=green!80!black, very thick, dotted, rectangle, anchor=north west,
     minimum width=5.8cm,
     minimum height=2.05cm]
    (box) at ($(a) + (-0.5, 0.30)$) {};
    \node[text=green!80!black, rotate=-90, anchor=south] at (box.east) {precomputed};
\end{tikzpicture}
\vspace{-1em}
\end{figure}

The acquisition functions described above are based on optimistic estimates of
the latent function $f$ which implicitly trade off between exploiting the
posterior mean and exploring based on the uncertainty.  We instead follow the
approach described in \cite{HennigSchuler:2012} and aim to maximize the expected
gain of information on the posterior distribution of the global maximizer
$\xopt$.  In Section~\ref{sec:method} we derive a rearrangement of this
acquisition function and a corresponding approximation that we call Predictive
Entropy Search (PES). PES is more accurate than the approximation used in
\cite{HennigSchuler:2012}.  In Section~\ref{sec:experiments} we empirically
evaluate this claim on both synthetic and real-world problems and show that this
leads to real gains in performance.

\section{Predictive entropy search}
\label{sec:method}

We propose to follow the information-theoretic method for active
data collection described in \cite{MacKay:1992}. We are interested in maximizing
information about the location $\xopt$ of the global maximum, whose posterior distribution
is $p(\xopt|\D_n)$. Our current information about $\xopt$ can
be measured in terms of the negative differential entropy of $p(\xopt|\D_n)$.
Therefore, our strategy is to select $\x_{n+1}$ which maximizes the expected
reduction in this quantity. The corresponding acquisition function is
\begin{equation}
    \textstyle
    \alpha_n(\x)
    = 
    \H[p(\xopt |\D_n)] -
    \E_{p(y|\D_n,\x)} [
        \H[p(\xopt|\D_n\cup\{(\x,y)\})]
    ]
    \,,
    \label{eq:originalObjective}
\end{equation}
where $\H[p(\x)] = -\int p(\x) \log p(\x) d\x$ represents the differential
entropy of its argument and the expectation above is taken with respect to
the posterior predictive distribution of $y$ given $\x$. 
The exact evaluation of (\ref{eq:originalObjective}) is infeasible in practice. The
main difficulties are i) $p(\xopt|\D_n\cup\{(\x,y)\})$ must be computed for
many different values of $\x$ and $y$ during the optimization of (\ref{eq:originalObjective}) and ii) the
entropy computations themselves are not analytical.  In practice, a direct
evaluation of (\ref{eq:originalObjective}) is only possible after performing
many approximations \cite{HennigSchuler:2012}.  To avoid this, we follow
the approach described in \cite{Houlsby:2012} by noting that
(\ref{eq:originalObjective}) can be equivalently written as the mutual
information between $\xopt$ and $y$ given $\D_n$. Since the mutual
information is a symmetric function, $\alpha_n(\x)$ can be rewritten as
\begin{equation}
    \textstyle
    \alpha_n(\x)
    = 
    \H[p(y|\D_n,\x)] -
    \E_{p(\xopt|\D_n)} [
        \H[p(y|\D_n,\x,\xopt)]
    ]
    \,,
    \label{eq:newObjective}
\end{equation}
where $p(y|\D_n,\x,\xopt)$ is the posterior predictive distribution for $y$
given the observed data $\D_n$ and the location of the global maximizer of $f$.
Intuitively, conditioning on the location $\xopt$ pushes the posterior predictions up in
locations around $\xopt$ and down in regions away from $\xopt$.
Note that, unlike the previous formulation, this objective is based on the
entropies of predictive distributions, which are analytic or can be easily
approximated, rather than on the entropies of distributions on $\xopt$ whose
approximation is more challenging.

The first term in (\ref{eq:newObjective}) can be computed
analytically using the posterior marginals for $f(\x)$ in (\ref{eq:gpvar}), that is,
$\H[p(y|\D_n,\x)]=0.5\log[2\pi e\,(v_n(\x)+\sigma^2)]$,
where we add $\sigma^2$ to $v_n(\x)$ because 
$y$ is obtained by adding Gaussian noise with variance $\sigma^2$ to $f(\mathbf{x})$.
The second term, on the other hand, must be approximated.
We first approximate the expectation in (\ref{eq:newObjective}) 
by averaging over samples $\xopt^{(i)}$ drawn approximately from $p(\xopt|\D_n)$.
For each of these samples, we then approximate the corresponding entropy function $\H[p(y|\D_n,\x,\xopt^{(i)})]$
using expectation propagation \cite{Minka:2001}. The code for all these operations is
publicly available at \url{http://tobediscloseduponacceptance.org}.

\subsection{Sampling from the posterior over global maxima}
\label{sec:sampling}

In this section we show how to approximately sample from the conditional
distribution of the global maximizer $\xopt$ given the observed data $\D_n$,
that is,
\begin{align}
    p(\xopt|\D_n) =
    p\big(f(\xopt) = \max_{\x\in\X} f(\x)\big|\D_n\big)\,.
\end{align}
If the domain $\X$ is restricted to some finite set of $m$ points,
the latent function $f$ takes the form of an $m$-dimensional vector $\mathbf{f}$.
The probability that the $i$th element of $\mathbf{f}$ is optimal can then be written
as $\int p(\mathbf{f}|\D_n) \prod_{j\leq m} \I[f_i \geq f_j] \,d\mathbf{f}$. This suggests the
following generative process: i) draw a sample from the posterior distribution $p(\mathbf{f}|\D_n)$ and
ii) return the index of the maximum element in the sampled vector. 
This process is known as Thompson sampling or probability matching 
when used as an arm-selection strategy in multi-armed bandits \cite{Li:2011}.
This same approach could be used for sampling the maximizer over a continuous
domain $\X$. At first glance this would require constructing an
infinite-dimensional object representing the function $f$.
To avoid this, one could sequentially construct $f$ while it is being optimized. However, evaluating
such an $f$ would ultimately have cost $\calO(m^3)$ where $m$ is the
number of function evaluations necessary to find the optimum.  
Instead, we propose to sample and optimize an analytic approximation to $f$.
We will briefly derive this approximation below, but more detail is given in Appendix~\ref{sec:samplingAppendix}.

Given a shift-invariant kernel $k$, Bochner's theorem~\citep{Bochner:1959}
asserts the existence of its Fourier dual $s(\vw)$, which is equal to the
spectral density of $k$.  Letting $p(\vw)=s(\vw)/\alpha$ be the associated normalized
density, we can write the kernel as the expectation
\begin{equation}
    k(\x,\x')
    = \alpha \,\E_{p(\vw)}[e^{-i\vw\T(\x-\x')}]
    = 2\alpha \,\E_{p(\vw,b)}[\cos(\vw\T\vx+b)\cos(\vw\T\vx' + b)]\,,
    \label{eq:kernel_approx}
\end{equation}
where $b\sim\Uniform[0,2\pi]$. Let $\vphi(\vx)=\sqrt{2\alpha/m} \cos(\vW\x +
\vb)$ denote an $m$-dimensional feature mapping where $\vW$ and $\vb$ consist
of $m$ stacked samples from $p(\vw,b)$. The
kernel $k$ can then be approximated by the inner product of these features,
$k(\x,\x')\approx \vphi(\x)\T\vphi(\x')$. This approach was used
by~\cite{Rahimi:2007} as an approximation method in the context of kernel
methods.  The feature mapping $\vphi(\x)$ allows us to approximate the Gaussian
process prior for $f$ with a linear model $f(\x)=\vphi(\vx)\T\vtheta$ where
$\vtheta\sim\Normal(\zero, \vI)$ is a standard Gaussian.  By conditioning on
$\D_n$, the posterior for $\vtheta$ is also multivariate Gaussian,
$\vtheta|\D_n\sim\Normal(\vA^{-1}\vPhi\T \vy_n, \sigma^2\vA^{-1})$ where
$\vA=\vPhi\T\vPhi+\sigma^2\vI$ and $\vPhi\T=[\vphi(\x_1)\dots\vphi(\x_n)]$.

Let $\vphi\i$ and $\vtheta\i$ be a random set of
features and the corresponding posterior weights sampled both according to the generative process given
above. They can then be used to construct the function
$f\i(\x)=\vphi\i(\vx)\T\vtheta\i$, which is an approximate posterior sample of
$f$---albeit one with a finite parameterization. We can then maximize this
function to obtain $\xopt\i=\argmax_{\x\in\X} f\i(\vx)$, which is 
approximately distributed according to $p(\xopt|\D_n)$. Note that for early
iterations when $n < m$, we can efficiently sample $\vtheta\i$ with cost
$\calO(n^2m)$ using the method described in Appendix B.2 of \cite{Seeger:2008}.
This allows us to use a large number of features in $\vphi\i(\x)$.

\subsection{Approximating the predictive entropy}\label{sec:predictiveEntropy}

We now show how to approximate $\H[p(y|\D_n,\x,\xopt)]$ in
(\ref{eq:newObjective}). Note that we can write the argument to $\H$ in this expression as
$p(y|\D_n,\x,\xopt) = \int p(y|f(\x)) p(f(\x)|\D_n,\xopt)\,df(\x)$.  
Here $p(f(\x)|\D_n,\xopt)$ is the posterior distribution on $f(\x)$ given 
$\D_n$ and the location $\xopt$ of the global maximizer of $f$.
When the likelihood $p(y|f(\x))$ is Gaussian, we have that
$p(f(\x)|\D_n)$ is analytically tractable since it is the predictive distribution
of a Gaussian process.  However, by further conditioning on the location $\xopt$
of the global maximizer we are introducing additional constraints, namely that $f(\vz)\leq f(\xopt)$ for all $\vz\in\X$.
These constraints make $p(f(\x)|\D_n,\xopt)$ intractable.  To circumvent this
difficulty, we instead use the following simplified constraints:
\begin{enumerate}
    \item[C1.] \textbf{$\xopt$ is a local maximum.} This is achieved by letting $\nabla
    f(\xopt)=\zero$ and ensuring that $\nabla^{2}f(\xopt)$ is negative
    definite. We further assume that the non-diagonal elements of
    $\nabla^{2}f(\xopt)$, denoted by $\upper[\nabla^{2}f(\xopt)]$, are known,
    for example they could all be zero. This simplifies the negative-definite constraint.
    We denote by C1.1 the constraint given by $\nabla f(\xopt)=\zero$ and $\upper[\nabla^2 f(\xopt)] = \zero$.
    We denote by C1.2 the constraint that forces the elements of $\diag[\nabla^2 f(\xopt)]$ to be negative.

    \item[C2.] \textbf{$f(\xopt)$ is larger than past observations.} 
    We also assume that $f(\xopt)\geq f(\x_i)$ for all $i\leq n$.
    However, we only observe $f(\x_i)$ noisily via $y_i$.
    To avoid making inference on these latent function values, we approximate the above hard constraints
    with the soft constraint $f(\xopt) > y_\mathrm{max} + \epsilon$, where
    $\epsilon\sim\Normal(0,\sigma^2)$ and $y_\mathrm{max}$ is the largest $y_i$ seen so far.
    
    \item[C3.] \textbf{$f(\x)$ is smaller than $f(\xopt)$.} This simplified
    constraint only conditions on the given $\x$ rather than requiring
    $f(\xopt)\leq f(\vz)$ for all $\vz\in\X$.
\end{enumerate}
We incorporate these simplified constraints into $p(f(\x)|\D_n)$ to 
approximate $p(f(\x)|\D_n,\xopt)$.  This is achieved by multiplying
$p(f(\x)|\D_n)$ with specific factors that encode the above constraints. In what
follows we briefly show how to construct these factors; more detail is given in
Appendix~\ref{sec:predictiveEntropyAppendix}.

\newcommand\C[1]{\mathrm{C#1}}





Consider the latent variable 
    $\vz = [f(\xopt);\, \diag[\nabla^2f(\xopt)]]$.
To incorporate constraint C1.1 we can condition on the data and on the
``observations'' given by the constraints $\nabla f(\xopt)=\zero$ and
$\upper[\nabla^2 f(\xopt)]=\zero$.  Since $f$ is distributed according to a GP,
the joint distribution between $\vz$ and these observations is multivariate
Gaussian. The covariance between the noisy observations $\vy_n$ and the extra
noise-free derivative observations can easily be computed \cite{Solak:2003}. The
resulting conditional distribution is also multivariate Gaussian with mean
$\vm_0$ and covariance $\vV_0$. These computations are similar to those
performed in (\ref{eq:gpvar}). Constraints C1.2 and C2 can then be incorporated
by writing
\begin{equation}
    \textstyle
    p(\vz|\D_n,\C1,\C2) \propto
    \Phi_{\sigma^2}(f(\xopt) - y_\text{max})
    \Big[ \prod_{i=1}^{d} \I\big([\nabla^2 f(\xopt)]_{ii} \leq 0\big) \Big]
    \Normal(\vz|\vm_0,\vV_0)\,,\label{eq:conditional}
\end{equation}
where $\Phi_{\sigma^2}$ is the cdf of a zero-mean Gaussian distribution with
variance $\sigma^2$.  The first new factor in this expression guarantees that
$f(\xopt) > y_\text{max} + \epsilon$, where we have marginalized $\epsilon$ out,
and the second set of factors guarantees that the entries in $\diag[\nabla^2
f(\xopt)]$ are negative.

Later integrals that make use of $p(\vz|\D_n,\C1,\C2)$, however, will not admit
a closed-form expression. As a result we compute a Gaussian approximation
$q(\vz)$ to this distribution using Expectation Propagation (EP)
\cite{Minka:2001}.  The resulting algorithm is similar to the implementation of
EP for binary classification with Gaussian processes \cite{Rasmussen:2006}. EP
approximates each non-Gaussian factor in (\ref{eq:conditional}) with a Gaussian
factor whose mean and variance are $\widetilde{m}_i$ and $\widetilde{v}_i$,
respectively. The EP approximation can then be written as
    $q(\vz)
    \propto [ \prod_{i=1}^{d+1} \Normal(z_i|\widetilde m_i, \widetilde v_i)
    ] \Normal(\vz|\vm_0,\vV_0)$.
Note that these computations have so far not depended on $\x$, so we can compute
$\{\vm_0,\vV_0, \widetilde\vm, \widetilde\vv\}$ once and store them for later
use, where $\widetilde\vm = (\tilde{m}_1,\ldots,\tilde{m}_{d+1})$ and
$\widetilde\vv = (\tilde{v}_1,\ldots,\tilde{v}_{d+1})$.

We will now describe how to compute the predictive variance of some latent
function value $f(\x)$ given these constraints. Let $\vf=[f(\vx); f(\xopt)]$ be a
vector given by the concatenation of the values of the latent function at $\x$ and $\xopt$.
The joint distribution between $\vf$,  $\vz$, the evaluations $\vy_n$ collected so far and
the derivative ``observations'' $\nabla f(\xopt)=\zero$ and
$\upper[\nabla^2 f(\xopt)]=\zero$ is multivariate Gaussian.
Using $q(\vz)$, we then obtain the following approximation:
\begin{equation}
    \textstyle
    p(\vf|\D_n,\C1,\C2) \approx
    \int p(\vf|\vz,\D_n,\C{1.1}) \,q(\vz)\,d \vz =
    \Normal(\vf|\vm_\vf,\vV_\vf)\,.
\end{equation}
Implicitly we are assuming above that $\vf$ depends on our observations and
constraint C1.1, but is independent of C1.2 and C2 given $\vz$.  The
computations necessary to obtain $\vm_\vf$ and $\vV_\vf$ are similar to those used
above and in (\ref{eq:gpvar}).
The required quantities are similar to the ones used by EP to make predictions in the
Gaussian process binary classifier~\cite{Rasmussen:2006}.  We can then
incorporate C3 by multiplying $\Normal(\vf|\vm_\vf,\vV_\vf)$ with a factor that
guarantees $f(\x) < f(\xopt)$.  The predictive distribution for $f(\x)$ given
$\D_n$ and all the constraints can be approximated as
\begin{align}
    \textstyle
    p(f(\x)|\D_n,\C1,\C2,\C3) 
    \approx
    Z^{-1} \int \I( f_1 < f_2) \,\Normal(\vf|\vm_\vf,\vV_\vf) \,df_2\,,
    \label{eq:finalPredictive}
\end{align}
where $Z$ is a normalization constant.  The variance of the right hand size of
(\ref{eq:finalPredictive}) is given by
\begin{equation}
    v_n(\x|\xopt)
    = [\vV_\vf]_{1,1} -
    s^{-1}\beta(\beta + \alpha)
    \{  [\vV_\vf]_{1,1} - [\vV_\vf]_{1,2} \}^2\,,
\label{eq:predictiveVariance}
\end{equation}
where $s = [-1, 1]\T \vV_\vf [-1, 1]$, $\alpha = \mu / \sqrt{s}$, $\mu = [-1, 1]\T
\vm_\vf$, $\beta = \phi(\alpha) / \Phi(\alpha)$, and $\phi(\cdot)$ and $\Phi(\cdot)$
are the standard Gaussian density function and cdf, respectively.  By further
approximating (\ref{eq:finalPredictive}) by a Gaussian distribution with the
same mean and variance we can write the entropy as
$\H[p(y|\D_n,\x,\xopt)] \approx 0.5 \log [2 \pi e (v_n(\x|\xopt) +
\sigma^2)  ]$.

The computation of (\ref{eq:predictiveVariance}) can be numerically unstable
when $s$ is very close to zero.  This occurs when $[\vV_\vf]_{1,1}$ is very similar
to $[\vV_\vf]_{1,2}$.  To avoid these numerical problems, we multiply $[\vV_\vf]_{1,2}$
by the largest $0\leq \kappa \leq 1$ that guarantees that $s > 10^{-10}$. This
can be understood as slightly reducing the amount of dependence between $f(\x)$
and $f(\xopt)$ when $\x$ is very close to $\xopt$.  Finally, fixing
$\upper[\nabla^2 f(\xopt)]$ to be zero can also produce poor predictions when
the actual $f$ does not satisfy this constraint.  To avoid this, we instead fix this quantity to
$\upper[\nabla^2 f\i(\xopt)]$, where $f\i$ is the $i$th
sample function optimized in Section \ref{sec:sampling} to sample $\xopt\i$.

\subsection{Hyperparameter learning and the PES acquisition function}
\label{sec:hyper-learning}

We now show how the previous approximations are integrated to compute the acquisition function used by 
predictive entropy search (PES).
This acquisition function performs a formal treatment of the hyperparameters. Let $\vpsi$ denote
a vector of hyperparameters which includes any kernel parameters as well as the
noise variance $\sigma^2$. Let $p(\vpsi|\D_n)\propto p(\vpsi)
\,p(\D_n|\vpsi)$ denote the posterior distribution over these
parameters where $p(\vpsi)$ is a hyperprior  and $p(\D_n|\vpsi)$ is the GP marginal likelihood.
For a fully Bayesian treatment of $\vpsi$ we must marginalize the acquisition
function (\ref{eq:newObjective}) with respect to this posterior. 
The corresponding integral has no analytic expression and must be approximated using Monte
Carlo. This approach is also taken in \cite{Snoek:2012}.

We draw $M$ samples $\{\vpsi\i\}$ from $p(\vpsi|\D_n)$ using
slice sampling~\cite{Vanhatalo:2012}. Let $\xopt\i$ denote a sampled global maximizer
drawn from $p(\xopt|\D_n,\vpsi\i)$ as described in Section \ref{sec:sampling}.  Furthermore, let $v_n\i(\x)$ and $v_n\i(\x|\xopt\i)$ denote
the predictive variances computed as described in
Section~\ref{sec:predictiveEntropy} when the model hyperparameters are fixed to $\vpsi\i$. 
We then write the marginalized acquisition function as
\begin{align}
    \alpha_n(\x) 
    &=
    \textstyle
    \frac{1}{M}\sum_{i=1}^M 
    \left\{ 0.5\log[v_n\i(\x)+\sigma^2] - 0.5\log[v_n\i(\x|\xopt\i)+\sigma^2]\right\}
    \,.\label{eq:marginalizedObjective}
\end{align}
Note that PES is effectively marginalizing the original acquisition function
(\ref{eq:originalObjective}) over $p(\vpsi|\D_n)$. This is a significant
advantage with respect to other methods that optimize the same
information-theoretic acquisition function but do not marginalize over
the hyper-parameters. For example, the approach of
\cite{HennigSchuler:2012} approximates
(\ref{eq:originalObjective}) only for fixed $\vpsi$. The resulting approximation is
computationally very expensive and recomputing it to average over multiple
samples from $p(\vpsi|\D_n)$ is infeasible in practice.  

Algorithm~\ref{alg:pes} shows pseudo-code for computing the PES acquisition
function. Note that most of the computations necessary for evaluating (\ref{eq:marginalizedObjective}) can be done
independently of the input $\x$, as noted in the pseudo-code. This initial cost
is dominated by a matrix inversion necessary to pre-compute $\vV$ for each
hyperparameter sample. The resulting complexity is $\calO[M(n+d + d(d-1)/2)^3]$.
This cost can be reduced to $\calO[M(n+d)^3]$ by ignoring the 
derivative observations imposed on $\upper[\nabla^2 f(\xopt)]$ by constraint C1.1.
Nevertheless, in the problems that we consider $d$ is very small (less than 20).
After these precomputations are done, the evaluation of (\ref{eq:marginalizedObjective})
is $\calO[M(n+d + d(d-1)/2)]$.


\section{Experiments}
\label{sec:experiments}

In our experiments, we use Gaussian process priors for $f$ with squared-exponential
kernels $k(\x,\x')=\gamma^2\exp\{-0.5 \sum_i (x_i-x_i')^2/\ell_i^2\}$.
The corresponding spectral density is zero-mean
Gaussian with covariance given by $\diag([\ell_i^{-2}])$ and normalizing constant $\alpha=\gamma^2$.
The model hyperparameters are $\{\gamma,\ell_1,\dots,\ell_d,\sigma^2\}$. 
We use broad, uninformative Gamma hyperpriors. 

First, we analyze the accuracy of PES in the task of approximating the differential entropy
(\ref{eq:originalObjective}).  We compare the PES approximation
(\ref{eq:marginalizedObjective}),
with the approximation used by the entropy
search (ES) method \cite{HennigSchuler:2012}.  We also compare with the ground
truth for (\ref{eq:originalObjective}) obtained using a rejection sampling (RS)
algorithm based on (\ref{eq:newObjective}).  For this experiment we generate the
data $\D_n$ using an objective function $f$ sampled from the Gaussian
process prior. The domain $\X$ of $f$ is fixed to be $[0,1]^2$
and data are generated using $\gamma^2=1$, $\sigma^2=10^{-6}$, and
$\ell_i^2=0.1$. To compute (\ref{eq:marginalizedObjective}) we avoid
sampling the hyperparameters and use the known values directly.
We further fix $M=200$ and $m = 1000$.

The ground truth rejection sampling scheme works as follows.  First, $\X$ is discretized
using a uniform grid. The expectation with respect to $p(\xopt|\D_{n})$ in
(\ref{eq:newObjective}) is then approximated using sampling. For this, we sample
$\xopt$ by evaluating a random sample from $p(f|\D_n)$ on each grid cell and
then selecting the cell with highest value.  Given $\xopt$, we then approximate
$\H[p(y|\D_{n},\x,\xopt)]$ by rejection sampling. We draw samples from
$p(f|\D_n)$ and reject those whose corresponding grid cell with highest value is
not $\xopt$.  Finally, we approximate $\H[p(y|\D_{n},\x,\xopt)]$ by first,
adding zero-mean Gaussian noise with variance $\sigma^2$ to the the evaluations
at $\x$ of the functions not rejected during the previous step and
second, we estimate the differential entropy of the resulting samples using
kernels \cite{Ahmad:1976}.

\begin{figure}
\begin{center}
\includegraphics[width=0.9\linewidth]{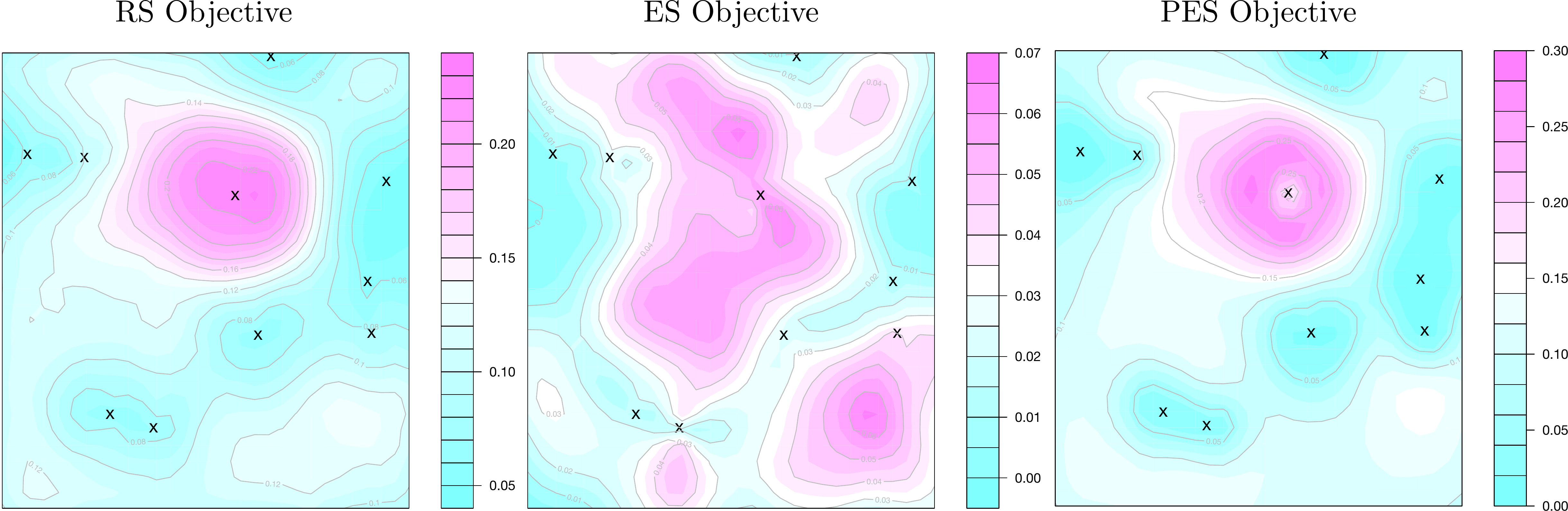}
\end{center}
\vspace{-0.5em}
\caption{\small Comparison of different estimates of the objective function $\alpha_n(\x)$
given by (\ref{eq:originalObjective}). Left, ground truth
obtained by the rejection sampling method RS.
Middle, approximation produced by the ES method. Right, approximation produced
by the proposed PES method. These plots show that the PES objective is much more
similar to the RS ground truth than the ES objective.}\label{fig:costFunctions}
\vspace{-0.5em}
\end{figure}

\begin{figure}
\begin{center}
\includegraphics[width=0.7\linewidth]{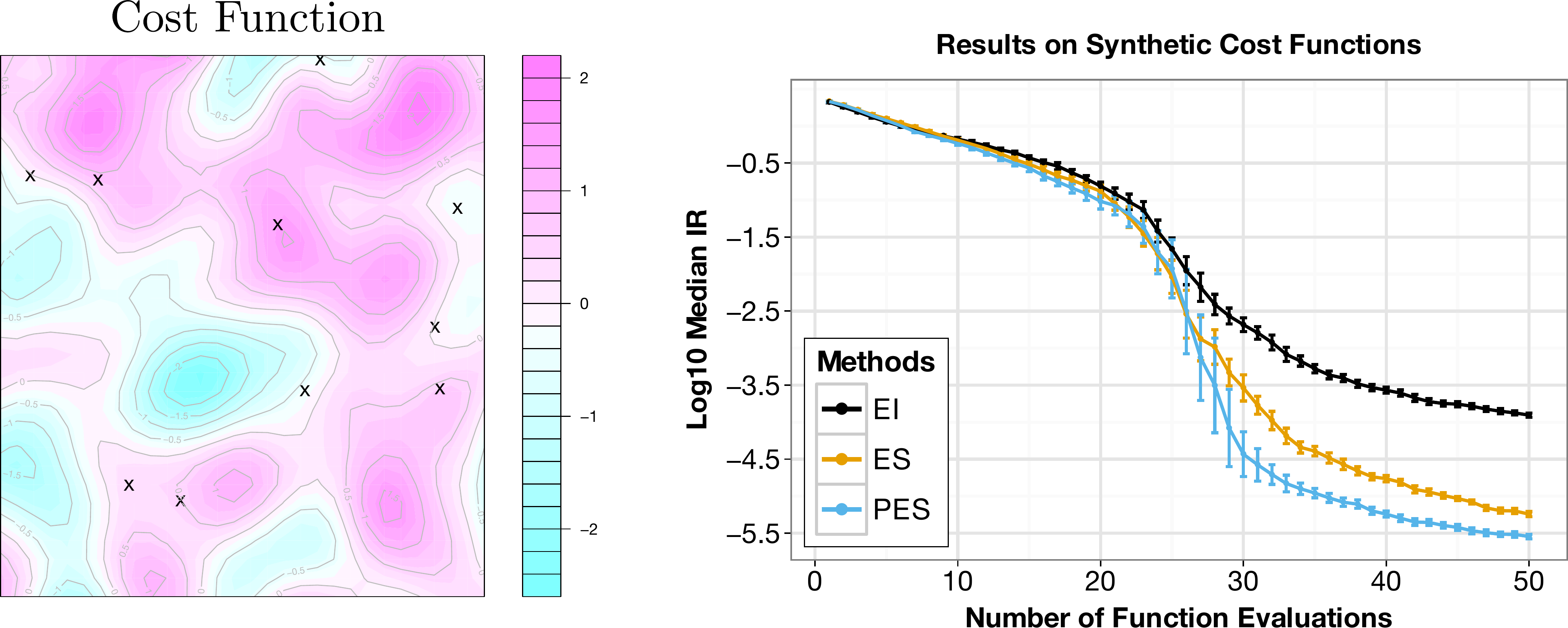}
\end{center}
\vspace{-1em}
\caption{\small Left, example of objective functions $f$. Right, median of the immediate
regret (IR) for the methods PES, ES and EI in the experiments with synthetic
objective functions.} \label{fig:resultsSyntheticFunctions}
\end{figure}

Figure~\ref{fig:costFunctions} shows the objective functions produced by RS, ES
and PES for a particular $\D_n$ with 10 measurements whose locations are
selected uniformly at random in $[0,1]^2$.  The locations of the collected
measurements are displayed with an ``x'' in the plots.  The particular objective function
used to generate the measurements in $\D_n$ is displayed in the left part of
Figure \ref{fig:resultsSyntheticFunctions}.  The plots in Figure
\ref{fig:costFunctions} show that the PES approximation to
(\ref{eq:originalObjective}) is more similar to the ground truth given by RS
than the approximation produced by ES.  In this figure we also see a
discrepancy between RS and PES at locations near $\x= (0.572, 0.687)$. This
difference is an artifact of the discretization used in RS. By
zooming in and drawing many more samples we would see the same behavior in both
plots.

We now evaluate the performance of PES in the task of finding the optimum of
synthetic black-box objective functions.  For this, we reproduce the within-model
comparison experiment described in \cite{HennigSchuler:2012}.  In this
experiment we optimize objective functions defined in the 2-dimensional unit domain
$\X = [0,1]^2$.  Each objective function is generated by first sampling 1024 function
values from the GP prior assumed by PES, using the same $\gamma^2$, $\ell_i$ and $\sigma^2$
as in the previous experiment.  The objective function is then given by the resulting
GP posterior mean.  We generated a total of 1000 objective functions by following
this procedure.  The left plot in Figure \ref{fig:resultsSyntheticFunctions}
shows an example function.  

In these experiments we compared the performance of PES with that of ES
\cite{HennigSchuler:2012} and expected improvement (EI) \cite{Jones:1998}, a
widely used acquisition function in the Bayesian optimization literature.  We
again assume that the optimal hyper-parameter values are known to all methods.
Predictive performance is then measured in terms of the immediate regret (IR)
$|f(\xrec_n) - f(\xopt)|$, where $\xopt$ is the known location of the global
maximum and $\xrec_n$ is the recommendation of each algorithm had we stopped at
step $n$---for all methods this is given by the maximizer of the posterior mean.
The right plot in Figure \ref{fig:resultsSyntheticFunctions} shows the decimal
logarithm of the median of the IR obtained by each method across the 1000
\emph{different} objective functions.  Confidence bands equal to one standard
deviation are obtained using the bootstrap method.  Note that while averaging
these results is also interesting, corresponding to the expected performance
averaged over the prior, here we report the median IR because
the empirical distribution of IR values is very heavy-tailed.  In this case, the
median is more representative of the exact location of the bulk of the data.
These results indicate that the best method in this setting is PES, which
significantly outperforms ES and EI.  The plot also shows that in this case ES
is significantly better than EI.

We perform another series of experiments in which we optimize well-known
synthetic benchmark functions including a mixture of cosines
\cite{Anderson:2000} and Branin-Hoo (both functions defined in $[0,1]^2$) as well as the
Hartmann-6 (defined in $[0,1]^6$) \cite{Lizotte:2008}.  In all instances, we fix the
measurement noise to $\sigma^2 = 10^{-3}$.  For both PES and EI we marginalize
the hyperparameters $\vpsi$ using the approach described in Section
\ref{sec:hyper-learning}.  ES, by contrast, cannot average its approximation of
(\ref{eq:originalObjective}) over the posterior on $\vpsi$.  Instead, ES works
by fixing $\vpsi$ to an estimate of its posterior mean (obtained using slice
sampling) \cite{Vanhatalo:2012}.  To evaluate the gains produced by the fully
Bayesian treatment of $\vpsi$ in PES, we also compare with a version of PES
(PES-NB) which performs the same non-Bayesian (NB) treatment of $\vpsi$ as ES.
In PES-NB we use a single fixed hyperparameter as in previous sections with
value given by the posterior mean of $\vpsi$.  All the methods are initialized
with three random measurements collected using latin hypercube sampling
\cite{Brochu:2009}.

\begin{figure}
\begin{center}
\includegraphics[width=0.85\linewidth]{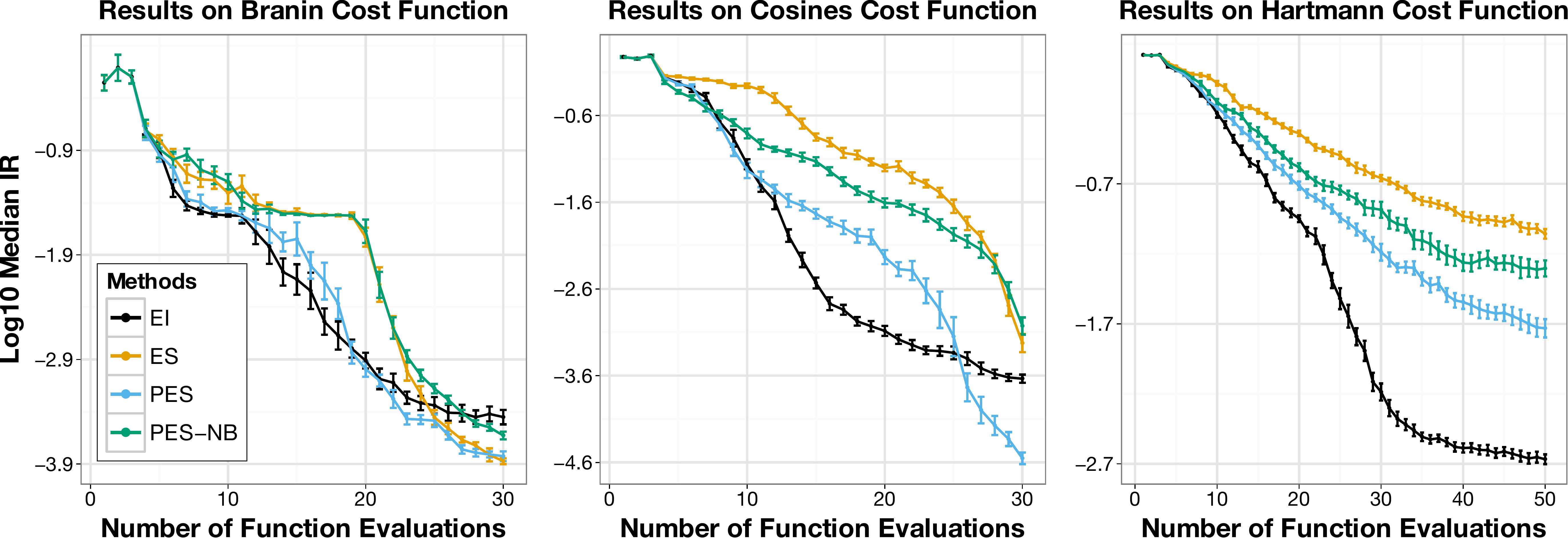}
\end{center}
\vspace{-0.5em}
\caption{\small Median of the immediate regret (IR) for the methods EI, ES, PES and
PES-NB in the experiments with well-known synthetic benchmark functions.}
\label{fig:resultsAnalyticRealWorldFunctions}
\vspace{-1.0em}
\end{figure}

The plots in Figure \ref{fig:resultsAnalyticRealWorldFunctions} show the median
IR obtained by each method on each function across 250 random initializations.
Overall, PES is better than PES-NB and ES. Furthermore, PES-NB is also
significantly better than ES in most of the cases.  These results show that the
fully Bayesian treatment of $\vpsi$ in PES is advantageous and that PES can
produce better approximations than ES.  Note that PES performs better than EI in
the Branin and cosines functions, while EI is significantly better on the
Hartmann problem. This appears to be due to the fact that entropy-based
strategies explore more aggressively which in higher-dimensional spaces takes
more iterations. The Hartmann problem, however, is a relatively simple problem
and as a result the comparatively more greedy behavior of EI does not result in
significant adverse consequences. Note that the synthetic functions optimized in the
previous experiment were much more multimodal that the ones considered here.

\subsection{Experiments with real-world functions}

We finally optimize different real-world cost functions. The 
first one (NNet) returns the predictive accuracy of a neural network on a random train/test partition of 
the Boston Housing dataset \cite{BacheLichman:2013}. The variables to optimize are the weight-decay parameter and the number of training
iterations for the neural network. The second function (Hydrogen) returns the amount of
hydrogen production of a particular bacteria in terms of the PH and Nitrogen levels of the growth medium \cite{Burrows:2009}.
The third one (Portfolio) returns the ratio of the mean and the standard deviation (the Sharpe ratio) of the 
1-year ahead returns generated by simulations from a multivariate time-series model that is adjusted to the daily returns of
stocks AXP, BA and HD. The time-series model is formed by univariate GARCH models connected with a Student's $t$ copula \cite{jondeau:2006}.
These three functions (NNet, Hydrogen and Portfolio) have as domain $[0,1]^2$. 
Furthermore, in these examples, the ground truth function that we want to optimize is unknown and is only available 
through noisy measurements. To obtain a ground truth, we approximate each cost function as the
predictive distribution of a GP that is adjusted to data sampled from the original function (1000 uniform samples for NNet and Portfolio
and all the available data for Hydrogen \cite{Burrows:2009}).
Finally, we also consider another real-world function that returns the walking
speed of a bipedal robot \cite{Westervelt:2007}.
This function is defined in $[0,1]^8$ and its inputs are the parameters
of the robot's controller. In this case the ground truth function is noiseless
and can be exactly evaluated through expensive numerical simulation. We consider two versions of this problem (Walker A) with zero-mean, additive noise of $\sigma=0.01$ and (Walker B) with $\sigma=0.1$.

\begin{figure}
\begin{center}
\includegraphics[width=1.00\linewidth]{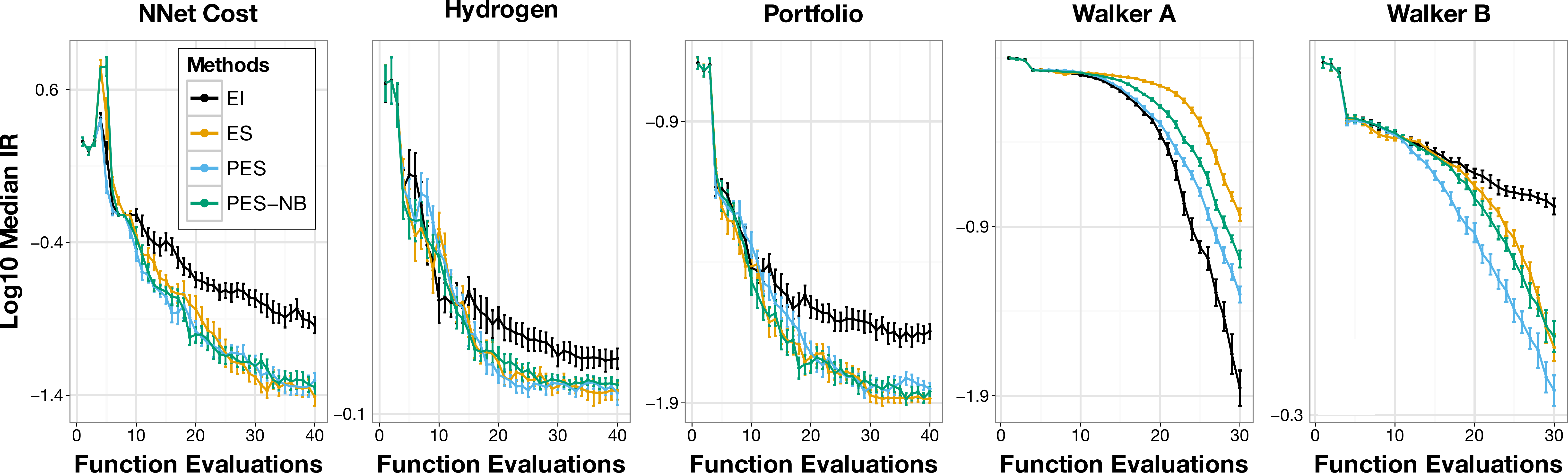}
\end{center}
\vspace{-0.5em}
\caption{\small Median of the immediate regret (IR) for the methods PES, PES-NB, ES and EI in the experiments with non-analytic
real-world cost functions.
}\label{fig:resultsNonAnalyticRealWorldFunctions}
\vspace{-0.5em}
\end{figure}

Figure \ref{fig:resultsNonAnalyticRealWorldFunctions} shows the median IR values
obtained by each method on each function across 250 random initializations, except in Hydrogen where we used 500 due to its higher level of noise.
Overall, PES, ES and PES-NB perform similarly in NNet, Hydrogen and Portfolio.
EI performs rather poorly in these first three functions. 
This method seems to make excessively greedy decisions and fails to explore the search space enough.
This strategy seems to be advantageous in Walker A, where EI obtains the best results.
By contrast, PES, ES and PES-NB tend to explore more in this latter dataset. This leads to worse results than those of EI.
Nevertheless, PES is significantly better than PES-NB and ES in both Walker datasets and better than EI in the noisier Walker B.
In this case, the fully Bayesian treatment of hyper-parameters performed by PES produces improvements in performance.

\section{Conclusions}

We have proposed a novel information-theoretic approach for Bayesian
optimization. Our method, predictive entropy search (PES), greedily maximizes
the amount of one-step information on the location $\xopt$ of the global maximum
using its posterior differential entropy.  Since this objective function is
intractable, PES approximates the original objective using a reparameterization
that measures entropy in the posterior predictive distribution of the function
evaluations.  PES produces more accurate approximations than Entropy Search
(ES), a method based on the original, non-transformed acquisition function.
Furthermore, PES can easily marginalize its approximation with respect to the
posterior distribution of its hyper-parameters, while ES cannot.  Experiments
with synthetic and real-world functions show that PES often outperforms ES in
terms of immediate regret. In these experiments, we also observe that PES often
produces better results than expected improvement (EI), a popular heuristic for
Bayesian optimization.  EI often seems to make excessively greedy decisions,
while PES tends to explore more. As a result, EI seems to perform better for
simple objective functions while often getting stuck with noisier objectives or
for functions with many modes.

\newpage
\addtolength{\bibsep}{-1pt}
{\small
\bibliographystyle{abbrv}
\bibliography{esbald}

\begin{thebibliography}{10}

\bibitem{Ahmad:1976}
I.~Ahmad and P.-E. Lin.
\newblock A nonparametric estimation of the entropy for absolutely continuous
  distributions.
\newblock {\em IEEE Transactions on Information Theory}, 22(3):372--375, 1976.

\bibitem{Anderson:2000}
B.~S. Anderson, A.~W. Moore, and D.~Cohn.
\newblock A nonparametric approach to noisy and costly optimization.
\newblock In {\em ICML}, pages 17--24, 2000.

\bibitem{BacheLichman:2013}
K.~Bache and M.~Lichman.
\newblock {UCI} machine learning repository, 2013.

\bibitem{Bochner:1959}
S.~Bochner.
\newblock {\em Lectures on Fourier integrals}.
\newblock Princeton University Press, 1959.

\bibitem{Brochu:2009}
E.~Brochu, V.~M. Cora, and N.~{de Freitas}.
\newblock A tutorial on {B}ayesian optimization of expensive cost functions,
  with application to active user modeling and hierarchical reinforcement
  learning.
\newblock Technical Report UBC TR-2009-23 and arXiv:1012.2599v1, Dept. of
  Computer Science, University of British Columbia, 2009.

\bibitem{Brochu:2007}
E.~Brochu, N.~{de Freitas}, and A.~Ghosh.
\newblock Active preference learning with discrete choice data.
\newblock In {\em NIPS}, pages 409--416, 2007.

\bibitem{Burrows:2009}
E.~H. Burrows, W.-K. Wong, X.~Fern, F.~W.~R. Chaplen, and R.~L. Ely.
\newblock Optimization of ph and nitrogen for enhanced hydrogen production by
  synechocystis sp. pcc 6803 via statistical and machine learning methods.
\newblock {\em Biotechnology Progress}, 25(4):1009--1017, 2009.

\bibitem{Li:2011}
O.~Chapelle and L.~Li.
\newblock An empirical evaluation of {Thompson} sampling.
\newblock In {\em NIPS}, pages 2249--2257, 2011.

\bibitem{HennigSchuler:2012}
P.~Hennig and C.~J. Schuler.
\newblock Entropy search for information-efficient global optimization.
\newblock {\em Journal of Machine Learning Research}, 13, 2012.

\bibitem{Hoffman:2011}
M.~W. Hoffman, E.~Brochu, and N.~de~Freitas.
\newblock Portfolio allocation for {Bayesian} optimization.
\newblock In {\em UAI}, pages 327--336, 2011.

\bibitem{Houlsby:2012}
N.~Houlsby, J.~M. Hern\'{a}ndez-lobato, F.~Huszar, and Z.~Ghahramani.
\newblock Collaborative gaussian processes for preference learning.
\newblock In {\em NIPS}, pages 2096--2104. 2012.

\bibitem{jondeau:2006}
E.~Jondeau and M.~Rockinger.
\newblock The copula-garch model of conditional dependencies: An international
  stock market application.
\newblock {\em Journal of international money and finance}, 25(5):827--853,
  2006.

\bibitem{Jones:1998}
D.~R. Jones, M.~Schonlau, and W.~J. Welch.
\newblock Efficient global optimization of expensive black-box functions.
\newblock {\em Journal of Global optimization}, 13(4):455--492, 1998.

\bibitem{Kushner:1964}
H.~Kushner.
\newblock A new method of locating the maximum of an arbitrary multipeak curve
  in the presence of noise.
\newblock {\em Journal of Basic Engineering}, 86, 1964.

\bibitem{Lizotte:2008}
D.~Lizotte.
\newblock {\em Practical {Bayesian} Optimization}.
\newblock PhD thesis, University of Alberta, Canada, 2008.

\bibitem{Lizotte:2007}
D.~Lizotte, T.~Wang, M.~Bowling, and D.~Schuurmans.
\newblock Automatic gait optimization with {Gaussian} process regression.
\newblock In {\em IJCAI}, pages 944--949, 2007.

\bibitem{MacKay:1992}
D.~J. MacKay.
\newblock Information-based objective functions for active data selection.
\newblock {\em Neural Computation}, 4(4):590--604, 1992.

\bibitem{Minka:2001}
T.~P. Minka.
\newblock {\em A family of algorithms for approximate Bayesian inference}.
\newblock PhD thesis, Massachusetts Institute of Technology, 2001.

\bibitem{Mockus:1978}
J.~Mo{\v c}kus, V.~Tiesis, and A.~{\v Z}ilinskas.
\newblock The application of bayesian methods for seeking the extremum.
\newblock In L.~Dixon and G.~Szego, editors, {\em Toward Global Optimization},
  volume~2. Elsevier, 1978.

\bibitem{negoescu:2011}
D.~M. Negoescu, P.~I. Frazier, and W.~B. Powell.
\newblock The knowledge-gradient algorithm for sequencing experiments in drug
  discovery.
\newblock {\em INFORMS Journal on Computing}, 23(3):346--363, 2011.

\bibitem{Rahimi:2007}
A.~Rahimi and B.~Recht.
\newblock Random features for large-scale kernel machines.
\newblock In {\em NIPS}, pages 1177--1184, 2007.

\bibitem{Rasmussen:2006}
C.~E. Rasmussen and C.~K. Williams.
\newblock {\em Gaussian processes for machine learning}.
\newblock The MIT Press, 2006.

\bibitem{Seeger:2008}
M.~W. Seeger.
\newblock Bayesian inference and optimal design for the sparse linear model.
\newblock {\em Journal of Machine Learning Research}, 9:759--813, 2008.

\bibitem{Snoek:2012}
J.~Snoek, H.~Larochelle, and R.~P. Adams.
\newblock Practical {Bayesian} optimization of machine learning algorithms.
\newblock In {\em NIPS}, pages 2960--2968, 2012.

\bibitem{Solak:2003}
E.~Solak, R.~Murray-smith, W.~E. Leithead, D.~J. Leith, and C.~E. Rasmussen.
\newblock Derivative observations in gaussian process models of dynamic
  systems.
\newblock In {\em NIPS}, pages 1057--1064. 2003.

\bibitem{Srinivas:2010}
N.~Srinivas, A.~Krause, S.~M. Kakade, and M.~Seeger.
\newblock Gaussian process optimization in the bandit setting: No regret and
  experimental design.
\newblock In {\em ICML}, pages 1015--1022, 2010.

\bibitem{Vanhatalo:2012}
J.~Vanhatalo, J.~Riihim{\"a}ki, J.~Hartikainen, P.~Jyl{\"a}nki, V.~Tolvanen,
  and A.~Vehtari.
\newblock Bayesian modeling with gaussian processes using the matlab toolbox
  gpstuff (v3.3).
\newblock {\em CoRR}, abs/1206.5754, 2012.

\bibitem{Wang:2013}
Z.~Wang, S.~Mohamed, and N.~de~Freitas.
\newblock Adaptive {Hamiltonian} and {Riemann Monte Carlo} samplers.
\newblock In {\em ICML}, 2013.

\bibitem{Westervelt:2007}
E.~Westervelt and J.~Grizzle.
\newblock {\em Feedback Control of Dynamic Bipedal Robot Locomotion}.
\newblock Control and Automation Series. CRC PressINC, 2007.

\end{thebibliography}
}

\newpage
\appendix
\section{Details on approximating GP sample paths}
\label{sec:samplingAppendix}

In this section we give further details about the approach used in
Section~\ref{sec:sampling} to approximate a GP using random features. 
These random features can be used to approximate sample paths from the GP
posterior. By optimizing these sample paths we obtain
posterior samples over the global maxima $\xopt$.  We derive in more
detail the kernel approximation from (\ref{eq:kernel_approx}). Formally, the
theorem of \citep{Bochner:1959} states
\begin{theorem}[Bochner's theorem]
    A continuous, shift-invariant kernel is positive definite if and only if it
    is the Fourier transform of a non-negative, finite measure.
\end{theorem}
\noindent As a result given some kernel $k(\x,\x')=k(\x-\x',\zero)$ there must
exist an associated density $s(\vw)$, known as its \emph{spectral density},
which is the Fourier dual of $k$. This can be written as
\begin{align*}
    k(\vx,\vx')
    &= \int e^{-i\vw\T(\vx-\vx')} s(\vw)\,d\vw,
    \\
    s(\vw)
    &= \frac1{(2\pi)^d} \int e^{i\vw\T\vtau} k(\vtau,\zero) \,d\vtau.
\end{align*}
Further, we can treat this measure as a probability density
$p(\vw)=s(\vw)/\alpha$ where $\alpha=\int s(\vw)\,d\vw$ is the normalizing
constant. Consequently, the kernel can be written as
\begin{align*}
    k(\vx,\vx')
    &= \alpha \,\E_{p(\vw)}[e^{-i\vw\T(\vx-\vx')}]
    \\
    \intertext{and due to the symmetry of $p(\vw)$ \citep[see][]{Rasmussen:2006}
    we can write the expectation as}
    &= \alpha \,\E_{p(\vw)}[\tfrac12 (e^{-i\vw\T(\vx-\vx')} + e^{i\vw\T(\vx-\vx')}) ]
    \\
    &= \alpha \,\E_{p(\vw)}[\cos(\vw\T\vx-\vw\T\vx')]
    \\
    \intertext{We can then note that $\int_0^{2\pi}\cos(a+2b)\,db=0$ for any
    constant offset $a\in\R$. As a result, for $b$ uniformly distributed between
    0 and $2\pi$ we can write}
    &= \alpha \,\E_{p(\vw)}[\cos(\vw\T\vx-\vw\T\vx') +
                \E_{p(  b)}[\cos(\vw\T\vx+\vw\T\vx'+2b)]]
    \\
    &= \alpha \,\E_{p(\vw,b)}[\cos(\vw\T\vx+b - \vw\T\vx' - b) +
                              \cos(\vw\T\vx+b + \vw\T\vx' + b)]
    \\
    &= 2\alpha \,\E_{p(\vw,b)}[\cos(\vw\T\vx+b)\cos(\vw\T\vx' + b)]
    \\
    \intertext{The last equality can be derived from the sum of angles formula,
    which leads to the identity: $2\cos(x)\cos(y)=\cos(x-y)+\cos(x+y)$. Finally,
    we can average over $m$ weights and phases}
    &=
    \frac{2\alpha}{m} 
    \,\E_{p(\vW,\vb)}[\cos(\vW\vx+\vb)\T\cos(\vW\vx' + \vb)]
\end{align*}
where $[\vW]_i\sim p(\vw)$ and $[\vb]_i\sim p(b)$ are stacked versions of the
original random variables. The resulting quantity has the same expectation but
results in a lower variance estimator. If we let
$\vphi(\vx)=\sqrt{2\alpha/m}\cos(\vW\vx+\vb)$ denote a random $m$-dimensional
feature generated by this model we can also write the kernel as
$k(\x,\x')=\E_{p(\vphi)}[\vphi(\vx)\T\vphi(\vx')]$.

We now briefly show the equivalence between a Bayesian linear model using random
features $\vphi$ and a GP with kernel $k$. Consider now a linear model
$f(\vx)=\phi(\vx)\T\vtheta$ where $\vtheta\sim\Normal(\zero,\vI)$ has a standard
Gaussian distribution and observations $\D_n=\{(\vx_i,y_i)\}_{i\leq n}$ of the
form $y_i\sim\Normal(f(\vx_i), \sigma^2)$. The posterior of $\vtheta$ given
$(\D_n,\vphi)$ is also be normal $\Normal(\vm,\vV)$ where
\begin{align*}
    \vm &= (\vPhi\T\vPhi + \sigma^2\vI)^{-1}\vPhi\T\vy, \\
    \vV &= (\vPhi\T\vPhi + \sigma^2\vI)^{-1}\sigma^2,
\end{align*}
and where $[\vPhi]_i=\vphi(\vx_i)$ and $[\vy]_i=y_i$ consist of the stacked
features and observations respectively. We can also easily write the predictive
distribution over $f$ evaluated at a test point $\vx$, which is Gaussian
distributed with mean and variance given by
\begin{align*}
    \mu_n(\vx)
    &= \vphi(\vx)\T(\vPhi\T\vPhi + \sigma^2\vI)^{-1} \vPhi\T \vy,
    \\
    v_n(\vx)
    &= \vphi(\vx)\T(\vPhi\T\vPhi + \sigma^2\vI)^{-1} \vphi(\vx) \sigma^2.
\end{align*}
By a simple application of the matrix-inversion lemma these quantities can be
rewritten in terms which only make use of the inner products between features,
\begin{align*}
    \mu_n(\vx)
    &= \vphi(\vx)\T \vPhi\T(\vPhi\vPhi\T + \sigma^2\vI)^{-1}\vy_{1:t}
    \\
    v_n(\vx)
    &= \vphi(\vx)\T\vphi(\vx)
     - \vphi(\vx)\T\vPhi\T(\vPhi\vPhi\T + \sigma^2\vI)^{-1}\vPhi\vphi(\vx).
\end{align*}
the expectations of which are equivalent to the kernel $k$ and we obtain the
same expressions as that in (\ref{eq:gpvar}).

\section{Details on approximating the predictive variance}
\label{sec:predictiveEntropyAppendix}

We now provide further details on approximating the predictive variance
$v_n(\x|\xopt)$ of inputs $\x$ given the position of the global optimizer
$\xopt$. In particular we include all steps omitted in the presentation of
Section~\ref{sec:predictiveEntropy}.

\subsection{Incorporating the analytic latent constraints (C1.1)}

We first turn to the random variables
\begin{align*}
    \vz &= [f(\xopt);\, \diag[\nabla^2f(\xopt)]], \\
    \vc &= [\vy_n; \nabla f(\xopt); \upper[\nabla^2 f(\xopt)]]
         = [\vy_n; \zero; \zero].
\end{align*}
Here $\vc$ contains the random variables that we will condition on in order to
enforce constraint C1.1. Given the input locations $\x$ and $\xopt$ we can
construct a kernel matrix $\vK$ containing the covariance evaluated on the
stacked vector $[\vz; \vc]$. We again refer to \cite{Solak:2003} in constructing
this matrix which includes derivative observations, the computations of which
are tedious but not overly complicated. Note also that the portions of $\vK$
which correspond to $y_i$ will have an additional $\sigma^2$ due to the
observation noise. Next let $\vK_\vz$, $\vK_\vc$, and $\vK_{\vz\vc}$ denote the
corresponding diagonal and off-diagonal blocks of the kernel matrix.
We can now condition on the observed values of $\vc$ to write
\begin{equation*}
    p(\vz|\D_n,\C{1.1})
    = p(\vz|\vc) = \Normal(\vz|\vm_0,\vV_0)
\end{equation*}
where $\vm_0=\vK_{\vz\vc} \vK^{-1}_\vc\vc$ and $\vV_0= \vK_\vz -
\vK_{\vz\vc}\vK^{-1}_\vc \vK_{\vz\vc}\T$.

\subsection{Incorporating the non-analytic latent constraints (C1.2 and C2)}

The additional constraints C1.2 and C2 can be introduced explicitly as in
(\ref{eq:conditional}), which takes the form of a single Gaussian factor and
$d+1$ non-Gaussian factors
\begin{equation*}
    p(\vz|\D_n,\C1,\C2)
    \propto
    \Normal(\vz|\vm_0,\vV_0)
    \Big[ \prod_{i=1}^{d+1} t_i(z_i) \Big].
\end{equation*}
We approximate this distribution using a single multivariate Gaussian $q(\vz)$
where each non-Gaussian factor is replaced by a Gaussian approximation
$\widetilde t_i(z_i)=\Normal(z_i;\widetilde m_i, \widetilde v_i)$ such that
\begin{equation*}
    q(\vz)
    = \Normal(\vz|\vm,\vV)
    \propto \Normal(\vz|\vm_0,\vV_0)
    \Big[ \prod_{i=1}^{d+1} \Normal(z_i;\widetilde m_i, \widetilde v_i) \Big]
\end{equation*}
where this approximation is parameterized by
$\vm=\vV[\widetilde\vV^{-1}\widetilde\vm+\vV_0^{-1}\vm_0]$ and
$\vV=(\widetilde\vV^{-1}+\vV_0^{-1})^{-1}$. The parameters of the approximate
factors are combined to form the vector $[\widetilde\vm]_i=\widetilde m_i$ and
the diagonal matrix $[\widetilde\vV]_{ii}=\widetilde v_i$. 

\def\oldm{\bar m_{i}}
\def\oldv{\bar v_{i}}
\def\norm{\bar Z_{i}}

To compute the approximate factors we use expectation propagation (EP). EP is a
procedure that starts from some initial values for the approximate factors
$(\widetilde m_i, \widetilde v_i)$ and iteratively refines these quantities;
here we initialize $\widetilde m_i=0$ and $\widetilde v_i=\infty$ which
corresponds to $\vm=\vm_0$ and $\vV=\vV_0$.  At each iteration, for every factor
$i$, we remove the contribution of the $i$th approximate factor to form the
\emph{cavity} distribution $q_{\setminus i}(\vz)\propto q(\vz)/\widetilde
t_i(z_i)$. Given the independent factors we consider here we can focus on each
individual component $q_{\setminus i}(z_i)$ separately with mean and variance
\begin{align*}
    \oldm &= \oldv(m_i/v_{ii} - \widetilde m_i/\widetilde v_i), \\
    \oldv &= (v_{ii}^{-1} - \widetilde v_i^{-1})^{-1}.
\end{align*}
Let $\hat q(z_i)\propto q_{\setminus i}(z_i) t_i(z_i)$ denote the \emph{tilted}
distribution where the $i$th approximate factor has been replaced by the
corresponding real factor. EP proceeds by finding the approximation $q_i$ that
minimizes the KL-divergence $\mathrm{D}[\hat q_i||q_i]$ where $q_i$ is
restricted to be Gaussian. This amounts to matching the first two moments.
Finally, by removing the influence of the cavity distribution and setting
$\widetilde t_i(z_i)\propto q_i(z_i)/q_{\setminus i}(z_i)$ we can update the
approximate factors. This can be performed using the same procedure which forms
the cavity distribution.

For both sets of constraints used in this work the moments can easily be
obtained by computing the normalizing constant $\norm = \int
\Normal(z_i|\oldm,\oldv) \,t_i(z_i) \,dz_i$ and using the following identities:
\begin{align*}
    \E_{\hat q}(z_i)
    &= \oldm + \oldv \frac{\partial\norm}{\partial\oldm},
    &
    \textrm{Var}_{\hat q}(z_i)
    &= \oldv - \oldv^2
    \left(
    \frac{\partial\norm}{\partial\oldm} 
    - 2\frac{\partial\norm}{\partial\oldv}\right).
\end{align*}
For the factors corresponding to constraints on the diagonal Hessian, i.e.\
where $t_i(z_i)=\I[z_i<0]$, the distribution is also simply a truncated
Gaussian.  Given these moments we can remove the contribution of the cavity
distribution as above and write
\begin{align*}
    \widetilde m_i &\gets \oldm - \kappa^{-1},
    &\text{where } 
      \alpha &= -\frac{\oldm}{\sqrt{\oldv}}, \\
    \widetilde v_i &\gets \beta^{-1} - \oldv,
    & \beta &=
      \frac{\phi(\alpha)}{\Phi(\alpha)}\left[\frac{\phi(\alpha)}{\Phi(\alpha)} +
      \alpha\right] \frac1{\oldv}, \\
    &&
    \kappa &= \left[\frac{\phi(\alpha)}{\Phi(\alpha)}-\alpha\right]\frac1{\sqrt{\oldv}}.
\end{align*}
For the final soft-maximum constraint,
$\Phi\big((z_i-y_\mathrm{max})/\sigma\big)$, the moments can be calculated in a
similar fashion. Using the same procedure as above we arrive at very similar
updates:
\begin{align*}
    \widetilde m_i &\gets \oldm + \kappa^{-1},
    &\text{where } 
      \alpha &= \frac{\oldm-y_\mathrm{max}}{\sqrt{\oldv+\sigma^2}}, \\
    \widetilde v_i &\gets \beta^{-1} - \oldv,
    & \beta &=
      \frac{\phi(\alpha)}{\Phi(\alpha)}\left[\frac{\phi(\alpha)}{\Phi(\alpha)} +
      \alpha\right] \frac1{\oldv+\sigma^2}, \\
    &&
    \kappa &= \left[\frac{\phi(\alpha)}{\Phi(\alpha)}+\alpha\right]\frac1{\sqrt{\oldv+\sigma^2}}.
\end{align*}

\subsection{Incorporating the prediction constraint (C3)}

Given some test input $\x$ we now turn to the problem of making predictions
about $f(\x)$. We again note that both the ``prior'' terms $\vm_0$, $\vV_0$ and
the EP factors, $\widetilde\vm$ and $\widetilde\vV$, are independent
of $\x$ and can be precomputed once for later use at prediction time. 

Let $\vf=[f(\vx); f(\xopt)]$ be a vector given by the concatenation of the
latent function at $\x$ and $\xopt$. The distribution for $\vf$ given the first
two constraints can be written as 
\begin{equation}
    p(\vf|\D_n,\C1,\C2)
    \approx
    \int p(\vf|\vz,\vc) \,q(\vz) \,d\vz = \Normal(\vf|\vm_\vf,\vV_\vf)\,.
\end{equation}
By writing $p(\vf|\vz,\vc)$ above we are assuming that $\vf$ is independent of
C1.2 and C2 given $\vz$ and as a result the above is simply an integral over the
product of two Gaussians. Let $\vK_\dagger$ be the cross-covariance matrix
evaluated between $\vf$ and $[\vz;\vc]$ and $\vK_\vf$ the covariance matrix
associated with $\vf$.  The posterior above will then be Gaussian with mean and
variance
\begin{align*}
    \vm_\vf &= \vK_\dagger [ \vK + \widetilde\vW]^{-1} [\vc; \widetilde\vm] \\
    \vV_\vf &= \vK_\vf - \vK_\dagger [\vK + \widetilde\vW]^{-1}\vK_\dagger\T,
\end{align*}
where $\widetilde\vW$ is a block-diagonal matrix where the first block is zero
and the second is $\widetilde\vV$ (note this matrix is also diagonal since
$\widetilde\vV$ is diagonal). Finally, these values can be plugged into
(\ref{eq:finalPredictive}--\ref{eq:predictiveVariance}) in order to arrive at
$v_n(\x|\xopt)$.

\end{document}